\documentclass[conference]{IEEEtran}
\IEEEoverridecommandlockouts
\usepackage{cite}
\usepackage{amsmath,amssymb,amsfonts}
\usepackage{algorithmic}
\usepackage{graphicx}
\usepackage{textcomp}
\usepackage{xcolor}
\usepackage[table]{xcolor}
\usepackage{marvosym}

\def\BibTeX{{\rm B\kern-.05em{\sc i\kern-.025em b}\kern-.08em
    T\kern-.1667em\lower.7ex\hbox{E}\kern-.125emX}}
\begin{document}

\title{CTIS-QA: Clinical Template-Informed Slide-level Question Answering for Pathology\\
% {\footnotesize \textsuperscript{*}Note: Sub-titles are not captured in Xplore and
% should not be used}
% \thanks{Identify applicable funding agency here. If none, delete this.}
}
\author{\IEEEauthorblockN{1\textsuperscript{st} Hao Lu*}
\IEEEauthorblockA{\textit{School of Biological Science and} \\
\textit{Medical Engineering} \\
\textit{Beihang University}\\
Beijing, China \\
lh2002@buaa.edu.cn}
\and
\IEEEauthorblockN{2\textsuperscript{nd} Ziniu Qian*}
\IEEEauthorblockA{\textit{School of Biological Science and} \\
\textit{Medical Engineering} \\
\textit{Beihang University}\\
Beijing, China \\
ziniuqian@buaa.edu.cn}
\and
\IEEEauthorblockN{3\textsuperscript{rd} Yifu Li}
\IEEEauthorblockA{\textit{School of Biological Science and} \\
\textit{Medical Engineering} \\
\textit{Beihang University}\\
Beijing, China \\
ferryliyifu@buaa.edu.cn}
\and
\IEEEauthorblockN{4\textsuperscript{th} Yang Zhou}
\IEEEauthorblockA{\textit{School of Biological Science and} \\
\textit{Medical Engineering} \\
\textit{Beihang University}\\
Beijing, China \\
zhouyangbme@buaa.edu.cn}
\and
\IEEEauthorblockN{5\textsuperscript{th} Bingzhen Wei}
\IEEEauthorblockA{\textit{ByteDance Inc.}\\
Beijing, China \\
bingzhengwei@hotmail.com}
\and
\IEEEauthorblockN{6\textsuperscript{th} Yan Xu \Letter}
\IEEEauthorblockA{\textit{School of Biological Science and} \\
\textit{Medical Engineering} \\
\textit{Beihang University}\\
Beijing, China \\
xuyan04@gmail.com}

\thanks{* indicates Equal Contribution. \Letter  indicates Corresponding Author}
\thanks{This work is supported by the National Natural Science Foundation in China under Grant 62371016 and U23B2063, the Bejing Natural Science Foundation Haidian District Joint Fund in China under Grant L222032, and the Academic Excellence Foundation of BUAA for PhD Students.} 
}
\maketitle

\begin{abstract}
Multimodal large language models (MLLMs) have demonstrated strong performance in patch-level pathological image analysis; however, they often lack the holistic perceptual capability necessary for comprehensive Whole Slide Image (WSI) interpretation. Recent approaches have explored constructing slide-level MLLMs using VQA datasets that are entirely generated from pathology reports by large language models (LLMs). However, these datasets suffer from critical limitations: hallucinated content, information leakage in question stems, clinically irrelevant or visual independent questions, and the omission of essential diagnostic features—issues that undermine both data quality and clinical validity. 
In this paper, we introduce a clinical diagnosis template-based pipeline to systematically collect and structure pathological information. In collaboration with pathologists and guided by the the College of American Pathologists (CAP) Cancer Protocols, we design a Clinical Pathology Report Template (CPRT) that ensures comprehensive and standardized extraction of diagnostic elements from pathology reports. We validate the effectiveness of our pipeline on TCGA-BRCA. First, we extract pathological features from reports using CPRT. These features are then used to build CTIS-Align, a dataset of 80k slide–description pairs from 804 WSIs for vision–language alignment training, and CTIS-Bench, a rigorously curated VQA benchmark comprising 977 WSIs and 14,879 question–answer pairs.
 CTIS-Bench emphasizes clinically grounded, closed-ended questions (e.g., tumor grade, receptor status) that reflect real diagnostic workflows, minimize non-visual reasoning, and require genuine slide understanding. 
We further propose \textbf{CTIS-QA}, a Slide-level Question Answering model, featuring a dual-stream architecture that mimics pathologists’ diagnostic approach. One stream captures global slide-level context via clustering-based feature aggregation, while the other focuses on salient local regions through attention-guided patch perception module. Extensive experiments on WSI-VQA, CTIS-Bench, and slide-level diagnostic tasks show that CTIS-QA consistently outperforms existing state-of-the-art models across multiple metrics. Code and data are available at https://github.com/HLSvois/CTIS-QA.
\end{abstract}


\section{Introduction}
\begin{figure}[!t]
    \centering
    \includegraphics[width=\columnwidth]{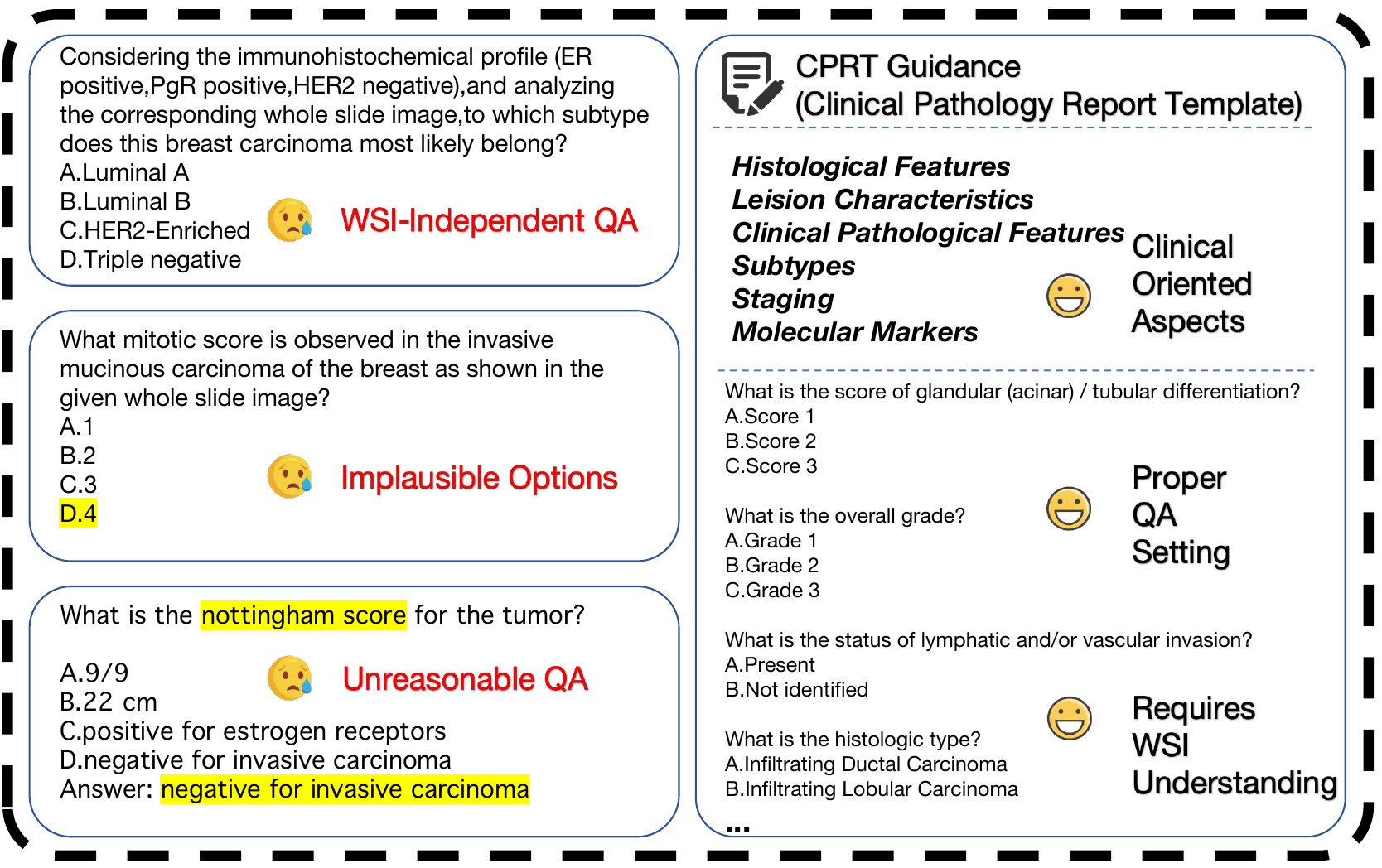} % 调整width比例
    \caption{Comparison between existing VQA datasets and CPRT-generated QA pairs. \textbf{Left:} Failure cases from existing VQA datasets. \textbf{Right:} CPRT-generated QA pairs demonstrating comprehensive coverage of essential pathological features following standardized clinical protocols, meeting the systematic requirements of real diagnostic workflows.}
    \label{fig:your_label}
\end{figure}
Whole Slide Images (WSIs) serve as the gold standard for clinical pathological diagnosis, encompassing comprehensive histomorphological information that is decisive for cancer diagnosis, staging, and treatment planning \cite{shao2021transmil,he2016deep}. However, WSI analysis requires detailed examination by experienced pathologists, demanding not only extensive specialized knowledge but also substantial time and human resources, presenting significant challenges in clinical practice due to the scarcity of expert resources and overwhelming workload \cite{chen2024wsi}.

To address this challenge, researchers have recently devoted efforts to developing artificial intelligence-based automated diagnostic models for WSIs, aimed at assisting clinicians in pathological analysis and generating standardized pathology reports \cite{seyfioglu2024quilt,sun2024pathgen,lu2023foundational,li2023llava,dai2024pa}. Pathology reports, as core clinical diagnostic documents, are enriched with critical pathological information contained within slides, including tumor grading, staging, immunohistochemical results, and other essential parameters required for clinical decision-making \cite{wu2023megacare,wu2025harnessing}. To train and evaluate these automated diagnostic models, constructing high-quality visual question answering datasets has become a crucial component, as the question-answering format enables systematic evaluation of models' understanding and recognition capabilities for various pathological features.

Some studies \cite{chen2024wsi,chen2024slidechat,liang2024wsi,sun2024cpath} have demonstrated the capability of Vision Language Models (VLMs) to utilize comprehensive contextual information at the slide level. In existing research, dataset construction typically involves inputting pathology reports \cite{kefeli2024tcga} into large language models \cite{achiam2023gpt,bi2024deepseek,chiang2023vicuna} combined with carefully designed prompts to automatically generate a fixed number of question-answer pairs. Despite meticulous optimization of prompts, this approach still presents several critical issues as shown in Fig.1. First, the extraction of pathological features lacks systematicity and comprehensiveness. Autonomous model judgment may overlook clinically important information such as tumor subtypes, microenvironmental changes, or rare morphological features. These elements play crucial roles in precise classification, prognostic assessment, and therapeutic decision-making. Second, a considerable proportion of the generated question-answer pairs can be answered through textual reasoning or common sense without requiring visual information. This makes it difficult to effectively evaluate models' visual understanding capabilities. Third, the datasets contain numerous open-ended questions with limited clinical utility. In contrast, actual pathological diagnosis predominantly involves closed-ended questions with defined options, such as grading and receptor status. 
% Finally, current evaluation commonly relies on natural language processing metrics such as BLEU and ROUGE, which cannot accurately reflect the consistency between model outputs and clinical diagnostic standards, making it challenging to assess their actual efficacy in real clinical scenarios.

To address these challenges, we propose a clinical diagnostic principles-guided pipeline for the systematic extraction of high-quality, clinically relevant question-answer pairs. Our approach is built upon a generalizable Clinical Pathology Report Template (CPRT) based framework grounded in the College of American Pathologists (CAP) Cancer Protocols \cite{sluijter2019improvement}, which provides standardized, evidence-based guidelines for comprehensive cancer reporting. This universal framework organizes pathological assessment into six core dimensions applicable across cancer types: Histological Features, Lesion Characteristics, Clinical Pathological Features, Subtypes, Staging, and Molecular Markers. By design, this structure ensures consistent and complete representation of diagnostically critical information, reducing omissions common in autonomous model-driven extraction.

In this work, we instantiate the general CPRT-based framework in the context of breast invasive carcinoma (BRCA), leveraging TCGA-BRCA pathology reports \cite{tomczak2015review}. Through close collaboration with pathologists, we adapt the template to capture essential disease-specific features while preserving the integrity of clinical diagnostic logic. This principled, standards-aligned approach ensures that the resulting question-answer pairs are not only visually grounded but also reflect the depth and rigor of real-world pathological evaluation, thereby enhancing their clinical relevance and diagnostic utility. Building on this pipeline, we introduce CTIS-Align, a dataset of 80k slide-description pairs derived from 804 WSIs, designed to enhance the vision–language alignment stage for MLLMs. We also propose CTIS-Bench, a Clinical Template-Informed Slide-level Benchmark featuring both closed-ended and open-ended clinically meaningful questions (e.g., tumor grade, receptor status) that reflect real diagnostic workflows. This design minimizes non-visual or clinically irrelevant QA pairs, ensuring that effective performance requires genuine WSI understanding rather than reliance on textual reasoning or prior knowledge, by aligning question formulation with actual clinical reporting standards.

% Furthermore, the benchmark is designed to support evaluation beyond conventional NLP metrics, enabling assessment of diagnostic consistency with clinical standards through structured output matching and expert-informed scoring criteria.

We propose \textbf{CTIS-QA}, a Clinical Template-Informed Slide-level Question Answering model based on the LLaVA framework \cite{liu2023visual}, designed for comprehensive understanding of Whole Slide Images (WSIs). Inspired by the diagnostic reasoning of pathologists—who integrate macroscopic tissue patterns across the entire slide with microscopic observations of localized regions—CTIS-QA employs a dual-stream architecture to effectively capture both global contextual information and fine-grained local features. The global stream uses a clustering-based mechanism to derive holistic slide-level representations, while the local stream leverages an attention module to selectively aggregate discriminative patch-level details. The fusion of these complementary visual pathways enables the model to perform fine-grained visual analysis grounded in whole-slide context, directly addressing the need for robust visual understanding in VQA tasks. The contributions of this work are three-fold:
\begin{enumerate}
    \item We propose a clinical diagnostic principles-guided pipeline with Clinical Pathology Report Template (CPRT) compliant with the CAP Cancer Protocols that ensures comprehensive and systematic extraction of clinically essential features.
    \item We propose CTIS-Align, a dataset of 80k slide–description pairs enabling robust vision–language alignment, and CTIS-Bench, a clinically informed benchmark with 977 WSIs and 14,879 question–answer pairs, featuring questions aligned with real diagnostic workflows, ensuring genuine visual reasoning rather than reliance on textual context alone.
    \item We propose CTIS-QA, a Clinical Template-Informed Slide-level Question Answering model that mimics pathologists' diagnostic workflow through a dual-stream architecture, enabling integrated reasoning over global tissue architecture and local morphological details for robust WSI understanding. 
\end{enumerate}

\section{Data Construction}
\begin{figure*}[!t]
    \centering
    \includegraphics[width=0.95\textwidth]{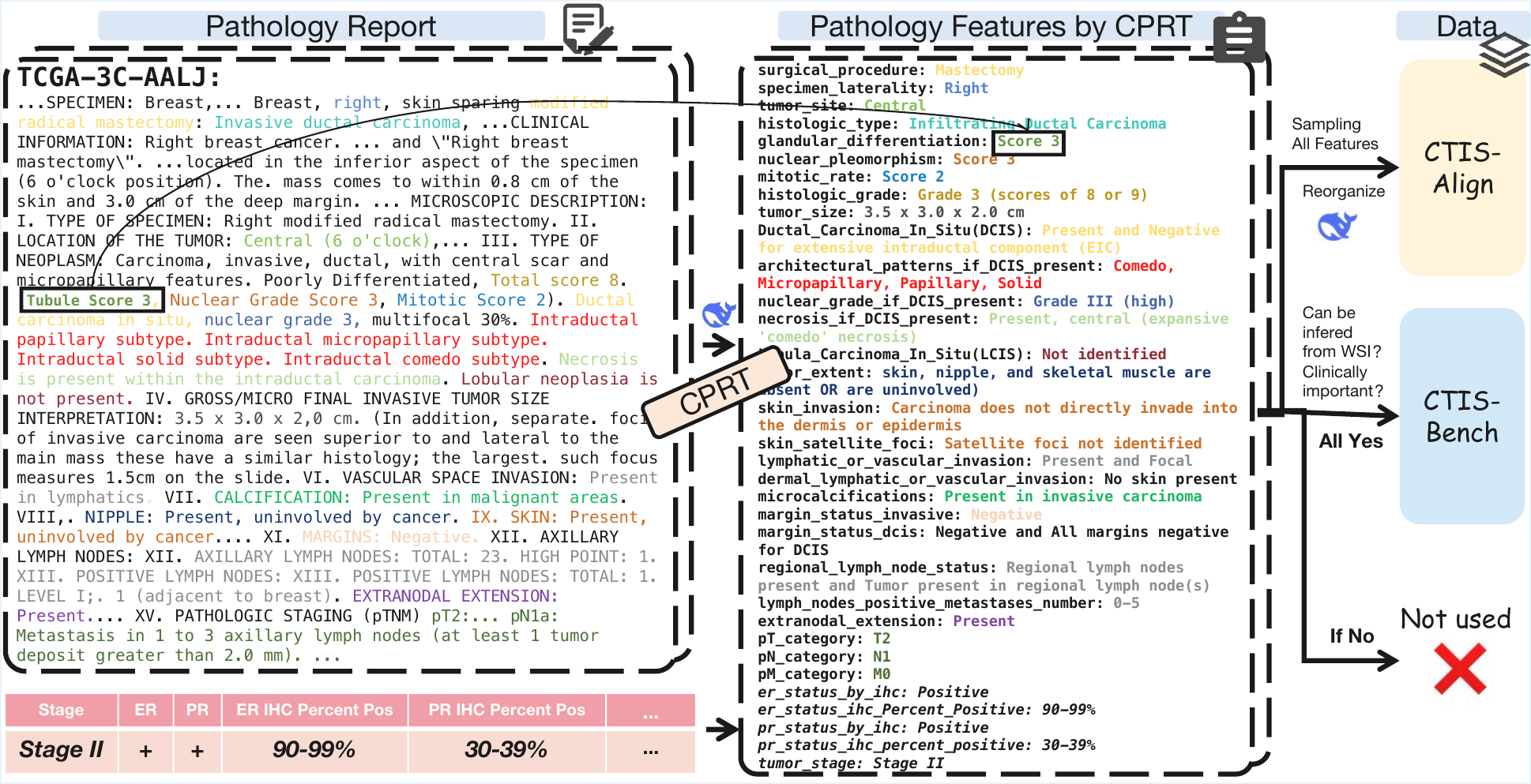} % 调整width比例
    \caption{Illustration of Data Construction Pipeline. Original pathology reports and metadata are processed through the Clinical Pathology Report Template (CPRT) with LLM assistance to extract comprehensive pathology features.  These features undergo random sampling and LLM reorganization to generate CTIS-Align for model pre-training.  Pathologist perform manual filtering and quality control to produce the final CTIS-Bench benchmark, ensuring clinical accuracy and diagnostic relevance.}
    \label{fig:your_label}
\end{figure*}
Our data construction pipeline processes original pathology reports and metadata on TCGA through the Clinical Pathology Report Template (CPRT) with LLM assistance to extract comprehensive pathology features as shown in Fig.2. These features undergo random sampling and LLM reorganization to generate CTIS-Align for model pre-training, while pathologists perform manual filtering and quality control to produce the final CTIS-Bench benchmark, ensuring clinical accuracy and diagnostic relevance. The statistics of CTIS-Align and CTIS-Bench are shown in Fig.3.

\subsection{Clinical Pathology Report Template}
We propose a general-purpose pipeline for CPRT construction, grounded in the College of American Pathologists (CAP) Cancer Protocols. These evidence-based guidelines provide a standardized framework for synoptic reporting across cancer types, ensuring comprehensive documentation of critical diagnostic elements and improving clinical decision-making, treatment planning, and reporting consistency. Our pipeline systematically organizes pathological evaluation into six universal dimensions applicable to any cancer type: \textbf{Histological Features} assesses tissue architecture, cellular morphology, and structural patterns; \textbf{Lesion Characteristics} evaluates the identification of early, precancerous, or non-invasive lesions; \textbf{Clinical Pathological Features} captures disease progression indicators and treatment-relevant findings; \textbf{Subtypes} tests precise differentiation among molecular or histological subtypes critical for therapy selection; \textbf{Staging} measures accuracy in determining disease extent and severity; and \textbf{Molecular Markers} evaluates recognition of key biomarkers related to prognosis and targeted therapy. This structured, modular framework ensures consistent and comprehensive assessment across diverse cancer types.  In this study, we instantiate this framework by focusing on invasive breast cancer, using the TCGA-BRCA cohort—a large and publicly available dataset with WSIs and paired pathology reports.  Leveraging this resource, we collaborate with clinical pathologists to develop a specialized CPRT that systematically incorporates 38 essential diagnostic elements, including histologic type and grade, DCIS status, margin assessment, lymph node involvement, hormone receptor (ER, PR) and HER2 status, and key pathological scoring systems.

The CPRT serves as a structured prompt framework to guide LLMs in generating comprehensive and clinically meaningful question-answer pairs, minimizing omissions of critical diagnostic information.  By grounding AI-driven analysis in established clinical standards, our approach mitigates the risk of overlooking subtle morphological features, tumor subtypes, or microenvironmental changes—issues commonly encountered in autonomous model interpretation.  This methodology transforms subjective visual assessments into a rigorous, standardized evaluation process, effectively bridging advanced AI capabilities with real-world pathology practice to support accurate diagnosis and informed therapeutic decisions in breast cancer management.
\begin{figure*}[t!]
    \centering
    \includegraphics[width=\textwidth]{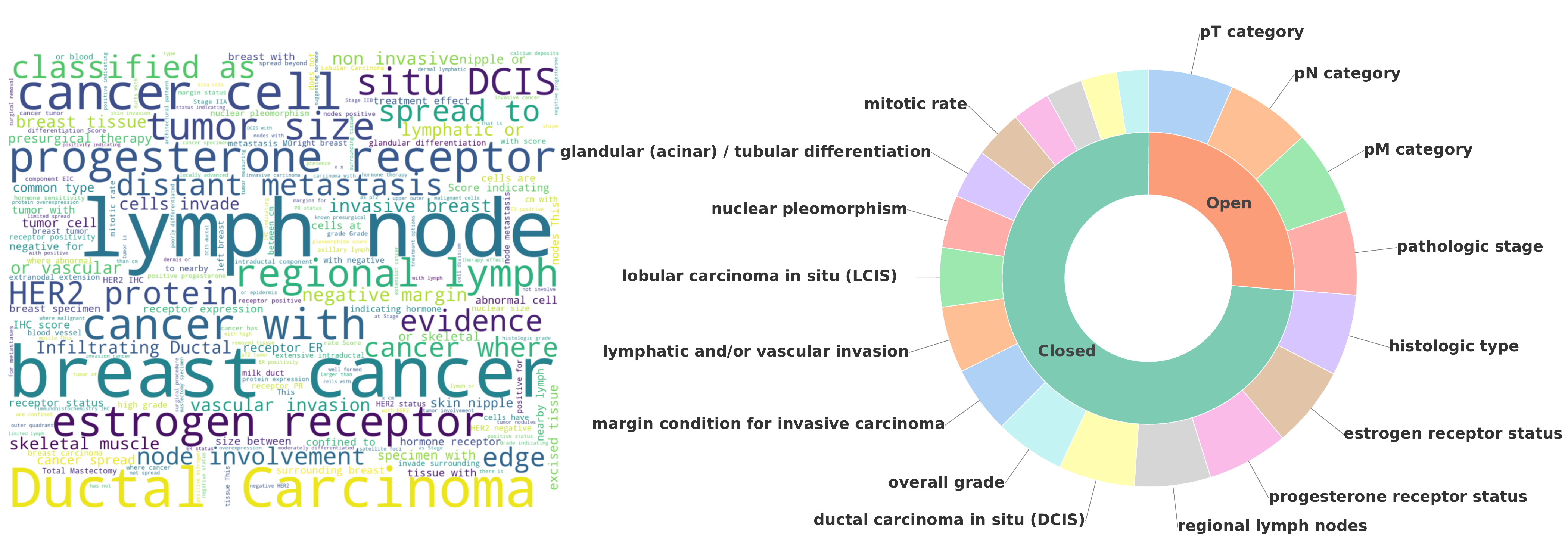} % 调整width比例
    \caption{Statistics of CTIS-Align and CTIS-Bench. The left part illustrates the word cloud of descriptions in CTIS-Align. The right part shows the question type distribution in the CTIS-Bench test set, the inner ring indicates whether the questions are open-ended or closed-ended.}
    \label{fig:your_label}
\end{figure*}
\subsection{Pathology Features Curation}
Building upon the CPRT developed for invasive breast cancer, we apply it to TCGA-BRCA to curate a comprehensive dataset. Using LLMs, we systematically extract all relevant diagnostic information from the pathology reports according to the CPRT structure. Each feature is extracted individually through specialized prompts that enforce alignment with predefined categorical options, ensuring consistency and clinical validity.
To ensure high accuracy, we implement a two-stage validation process. First, the same LLM performs self-verification by comparing the extracted key-value pairs against the original reports to detect and correct inconsistencies. Second, clinical pathologists conduct manual spot-checks on randomly sampled cases to validate the overall quality and reliability of the extractions. After rigorous extraction and validation, we obtain 22,435 high-quality key-value pairs from 977 WSIs, with an average of 22.96 pairs per slide covering essential pathological features. The dataset is split into training (804 WSIs), validation (87 WSIs), and test (86 WSIs) sets, preserving the distribution of pathological features across all subsets.

\subsection{CTIS-Align Dataset Construction}
In our method, we freeze the LLM and align WSI features with its semantic space using CTIS-Align dataset. Training directly with WSI-report pairs poses two significant challenges: limited data volume risking overfitting, and extensive text potentially causing the model to focus on report format rather than critical pathological information. To address the challenges of limited data volume and extensive text format in WSI-report pairs, we implement a systematic three-stage pipeline for constructing the CTIS-Align dataset from TCGA-BRCA pathological reports. First, we extract and standardize key pathological features from TCGA reports using our Clinical Pathology Report Template (CPRT), which was developed in collaboration with pathologists following CAP Cancer Protocols. This template-guided extraction ensures comprehensive coverage of clinically essential diagnostic elements while maintaining consistency across diverse report formats. Second, we employ AI-assisted validation to ensure feature consistency and accuracy across the dataset. Finally, we generate diverse descriptions through strategic feature combination, where each case contributes 100 description samples by randomly selecting 3-5 features per combination. The description generation utilizes professional pathologist-oriented prompt engineering to produce coherent sentences that integrate pathological explanations naturally. This systematic approach yields 80,000 high-quality QA pairs for CTIS-Align, providing both data diversity and pathological accuracy essential for effective WSI feature alignment training.

\begin{figure}[!h]
    \centering
    \includegraphics[width=0.9\columnwidth]{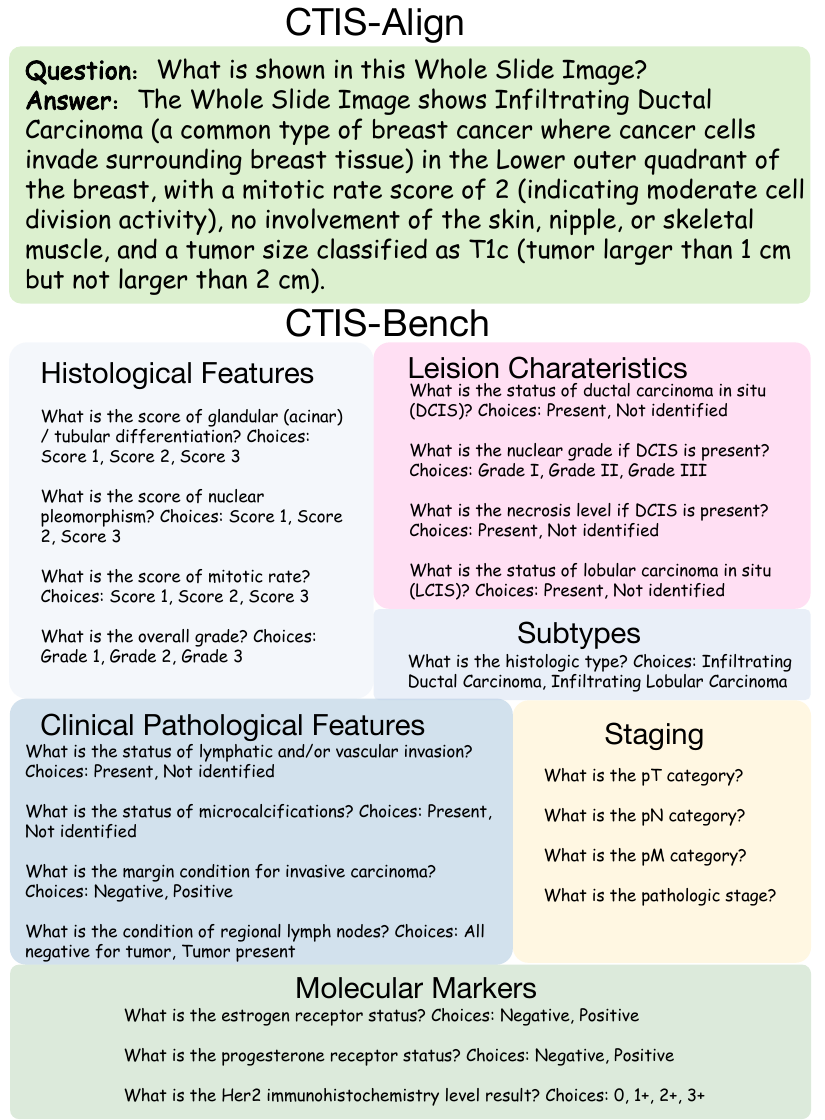} % 调整width比例
    \caption{Examples of CTIS-Align and CTIS-Bench datasets. CTIS-Align demonstrates a sampled and reorganized QA pair for WSI analysis. CTIS-Bench presents six key pathological aspects  along with their corresponding questions}
    \label{fig:your_label}
\end{figure}

\begin{figure*}[h]
    \centering
    \includegraphics[width=0.95\textwidth]{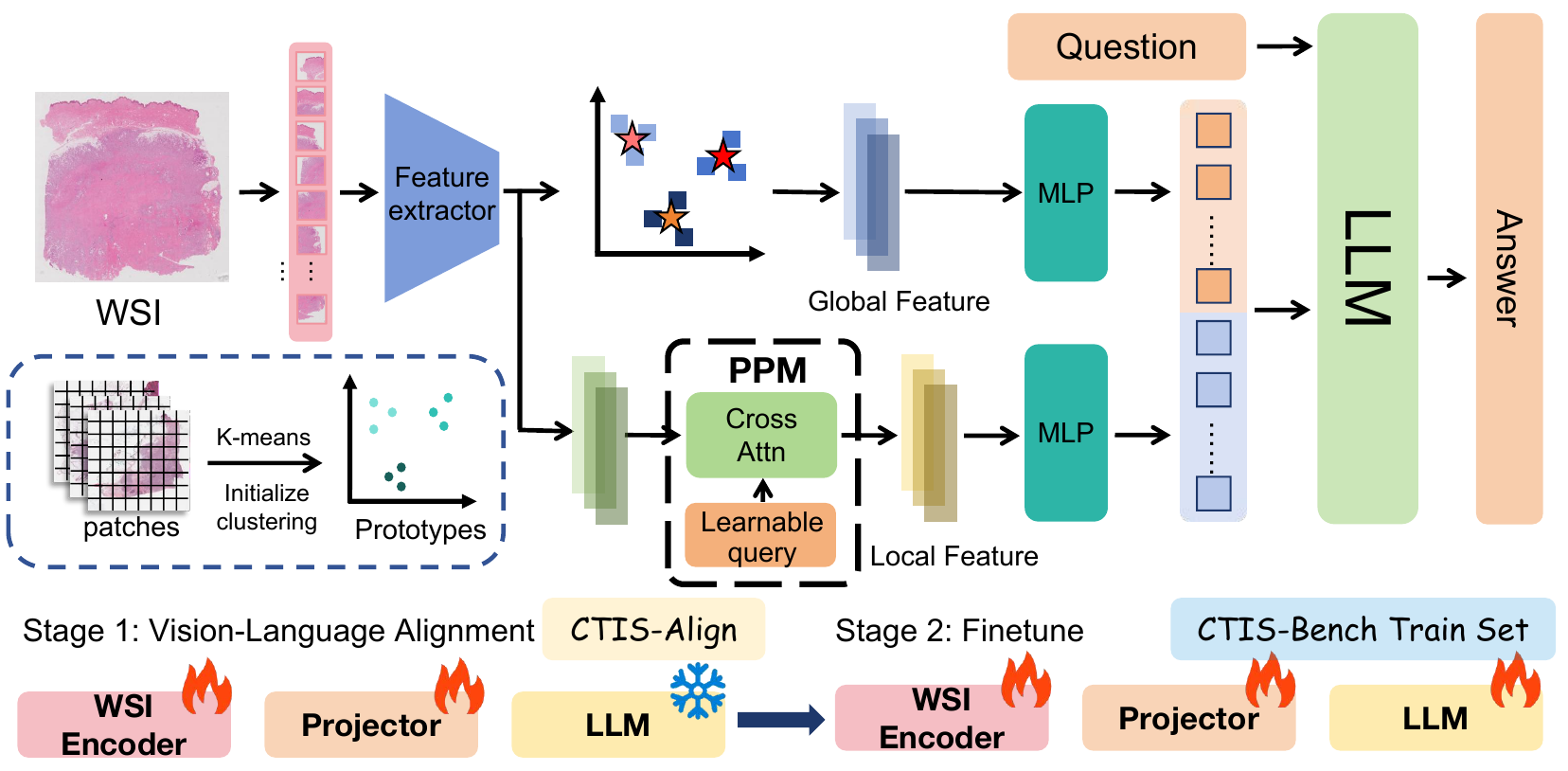} % 调整width比例
    \caption{Overall Structure of the CTIS-QA Model. Left bottom part shows the prototype initialize process. The lower section illustrates the two-stage training process. }
    \label{fig:your_label}
\end{figure*}

\subsection{CTIS-Bench Dataset Construction}
We construct the CTIS-Bench dataset based on previously obtained pathology features from 977 WSIs. To ensure both visual answerability and clinical relevance, pathologists carefully curated 38 candidate questions across the six core aspects of the CPRT framework. These questions were rigorously filtered according to two key criteria: (1) whether the answer can be confidently derived from the visual content of the WSI, and (2) whether the question holds significant diagnostic value in clinical practice. After this refinement process, we selected 20 high-quality, clinically meaningful questions that meet both criteria. The final dataset contains 14,879 question-answer pairs, each corresponding to one of the 20 standardized questions.

In Fig.4, we provide examples for CTIS-Align and CTIS-Bench.

\section{Method}

We propose CTIS-QA, a novel visual question answering framework specifically designed for WSI content understanding.  Unlike natural images, WSIs exhibit complex hierarchical structures requiring both comprehensive contextual understanding and precise detail extraction for accurate pathological diagnosis.

To tackle these challenges, our approach extends the LLaVA\cite{liu2023visual} architecture with a dual-stream visual encoder that synergistically captures complementary visual representations.  The global feature stream employs a clustering-based mechanism to extract slide-level semantic representations, providing essential contextual understanding of overall tissue organization and structural patterns crucial for answering questions about histological concepts and relationships.  Complementarily, the local feature stream leverages a multi-head attention mechanism to aggregate fine-grained visual details, enabling precise identification of specific morphological features, cellular components, and localized pathological elements essential for detail-oriented diagnostic questions.  The dual-stream design ensures comprehensive visual understanding by combining holistic slide-level insights with focused local analysis.

Our training follows a two-stage process within the LLaVA framework.  In the first stage, we use CTIS-Align to perform vision-language alignment training while freezing the LLM parameters, enabling the model to learn pathology-specific visual representations.  In the second stage, we fine-tune the entire model using the CTIS-Bench training set to optimize performance on clinical diagnostic tasks.  Following feature extraction, both streams are integrated through a learnable fusion module, with fused features projected into the language model's embedding space to maintain LLaVA compatibility while enhancing WSI-specific visual comprehension capabilities.  An overall framework of the proposed CTIS-QA is illustrated in Fig.5.

\subsection{Dual-stream Visual Encoder}

\textbf{Global Feature Stream.} To capture comprehensive slide-level contextual information, we design a clustering-based global feature extraction mechanism that aggregates semantic representations across the entire slide. We first divide the input slide image into $256 \times 256$ pixel patches at high resolution, ensuring sufficient granularity for feature extraction. These patches are processed by a frozen patch-level encoder to obtain feature vectors $\{ \mathbf{x}_i \}_{i=1}^N$, where $N$ represents the total number of patches extracted from the slide.

The global representation construction employs a K-means clustering approach to identify dominant visual patterns and thematic structures within the slide. We initialize $K$ prototypes $\{ \mathbf{c}_k \}_{k=1}^K$, where each $\mathbf{c}_k$ corresponds to the initial centroid of cluster $k$. 
The K-means clustering process minimizes the following objective function:
\begin{equation}
\mathbf{C}^* = \arg \min_{\mathbf{C}} \sum_{i=1}^N \sum_{k=1}^K \mathbf{1}(\mathbf{z}_i = k) \|\mathbf{x}_i - \mathbf{c}_k\|^2_2,
\end{equation}
where $\mathbf{C} = \{\mathbf{c}_k\}_{k=1}^K$ represents the set of centroids, $\mathbf{z}_i$ denotes the cluster assignment of patch $\mathbf{x}_i$, and $\mathbf{1}(\mathbf{z}_i = k)$ is the indicator function that equals 1 when patch $\mathbf{x}_i$ is assigned to centroid $\mathbf{c}_k$, and 0 otherwise. The term $\|\mathbf{x}_i - \mathbf{c}_k\|_2$ represents the Euclidean distance between patch feature $\mathbf{x}_i$ and centroid $\mathbf{c}_k$.

After convergence, the global representation aggregates information from all cluster centroids:
\begin{equation}
\mathbf{R}_{\text{global}} = \frac{1}{K} \sum_{k=1}^K \mathbf{c}_k,
\end{equation}
where $\mathbf{R}_{\text{global}}$ encodes the holistic semantic structure of the slide by averaging the learned centroids, capturing slide-level patterns such as layout organization, content distribution, and thematic coherence. Following \cite{song2024morphological}, we set the number of clusters to 16.

\textbf{Local Feature Stream.} Complementary to the global stream, our local feature stream employs a Patch Perception Module designed to capture fine-grained visual details and spatially-localized information. This module aggregates patch-level features through an attention-based mechanism that selectively focuses on the most relevant regions for question answering while preserving contextual relationships.

The local feature aggregation process utilizes multi-head self-attention to compute attention weights across all patch features. This mechanism enables the model to dynamically attend to specific visual elements such as textual content, graphical components, and detailed structural information that are crucial for answering precise questions about slide content, while maintaining awareness of spatial relationships and contextual dependencies among different slide regions.

% \begin{figure}[!t]
%     \centering
%     \includegraphics[width=\textwidth,trim=25 220 150 30,clip]{Fig2.pdf}
%     \caption{Overall Structure of the CTIS-QA Model. Left bottom part shows the prototype initialize process.}
%     \label{fig2}
% \end{figure}

\subsection{Patch Perception Module}
To preserve fine-grained morphological details that may be lost in global pooling operations, we introduce the Patch Perception Module (PPM), which enhances the model's sensitivity to diagnostically critical local structures. Given a WSI decomposed into $N$ patches with corresponding features $\{ \mathbf{x}_i \}_{i=1}^N$ from the patch-level encoder, we standardize the input length to ensure computational consistency. Let $M$ denote the maximum allowed patch count: if $N < M$, we pad the sequence with zero vectors; if $N > M$, we divide the patch list into $K$ equal segments and uniformly sample $M/K$ patches from each segment to maintain spatial distribution.

Let $X \in \mathbb{R}^{M \times d}$ denote the processed patch features after length normalization. The PPM employs a learnable query mechanism via cross-attention to extract salient local representations. Specifically, we define $L$ learnable query vectors $Lq \in \mathbb{R}^{L \times d_p}$, which are trained to attend to discriminative patch features. The local representation $\mathbf{R}_{\text{local}} \in \mathbb{R}^{L \times d}$ is computed as:

\begin{equation}
\mathbf{R}_{\text{local}} = \text{softmax}\left(\frac{W_q Lq (W_k X)^T}{\sqrt{d}}\right) W_v X,
\end{equation}

where $W_q \in \mathbb{R}^{d \times d}$, $W_k \in \mathbb{R}^{d \times d}$, and $W_v \in \mathbb{R}^{d \times d}$ are learnable projection matrices. This operation compresses the $M$ patch features into $L$ context-aware tokens that emphasize diagnostically relevant regions—such as mitotic figures or invasive fronts—while suppressing redundant background areas. The resulting $\mathbf{R}_{\text{local}}$ is then projected into the textual semantic space via a dedicated visual projector and concatenated with the global representation for holistic WSI understanding.

\begin{table*}[h!]
\centering
\caption{Comparison on WSI-VQA and CTIS-Bench benchmarks. WSI-VQA reports BLEU, METEOR, RougeL, and Accuracy; CTIS-Bench reports per-category accuracy and overall average. Since GPT-4o tends to generate excessively long responses for open-ended questions, metrics such as BLEU, METEOR, and Staging Accuracy cannot be reliably computed; we therefore denote these cases with “/” and omit its average metrics.}
\renewcommand{\arraystretch}{1.2}
\resizebox{\textwidth}{!}{%
\begin{tabular}{lcccccccccccc}
\hline
\textbf{Method} 
& \multicolumn{5}{c}{\textbf{WSI-VQA}} 
& \multicolumn{7}{c}{\textbf{CTIS-Bench}} \\
\hline
 & BLEU1 & BLEU4 & METEOR & RougeL & Acc 
 & H.F. & L.C. & C.P.F. & Subtypes & Staging & M.M. & Avg \\
\hline
GPT-4o 
 & / & / & / & / & 0.348 
 & 0.070 & 0.407 & 0.512 & 0.497 & / & 0.091 & / \\
Quilt-LLaVA 
 & 0.718 & 0.625 & 0.433 & 0.660 & 0.615 
 & 0.412 & 0.494 & 0.517 & 0.577 & 0.483 & 0.460 & 0.491 \\
WSI-VQA 
 & 0.383 & 0.167 & 0.223 & 0.455 & 0.492 
 & 0.459 & 0.568 & 0.555 & 0.740 & 0.503 & 0.602 & 0.571 \\
 \hline
 \rowcolor{gray!20}
CTIS-QA(Ours) 
 & \textbf{0.767} & \textbf{0.701} & \textbf{0.479} & \textbf{0.701} & \textbf{0.649} 
 & \textbf{0.523} & \textbf{0.635} & \textbf{0.599} & \textbf{0.851} & \textbf{0.541} & \textbf{0.629} & \textbf{0.630} \\
\hline
\end{tabular}%
}
\label{tab:wsi_slide_results}
\end{table*}

\section{Experiments}
\subsection{Datasets}
\textbf{WSI-VQA}: WSI-VQA \cite{chen2024wsi} is proposed by Chen et al.\cite{chen2024wsi}, which is build upon TCGA-BRCA dataset, comprising 977 pathology slides and 8,671 VQA pairs. Following the original protocol, the dataset is partitioned into training (804 slides/7,139 pairs), validation (87 slides/798 pairs), and test sets (86 slides/735 pairs). Evaluation metrics include BLEU-1 \cite{papineni2002bleu}, BLEU-4, METEOR \cite{banerjee2005meteor}, ROUGE-L \cite{lin2004rouge}, and closed-set accuracy.

\textbf{CTIS-Bench}: CTIS-Bench details are outlined in Section 2.3. We maintain identical patient splits as in WSI-VQA, with 12256/1319/1304 QA pairs in the training/validation/test sets, respectively. We employ balanced accuracy as the primary evaluation metric of six aspect: Histological Features (H.F.), Lesion Characteristics (L.C.), Clinical Pathological Features (C.P.F.), Subtypes, Staging and Molecular Markers (M.M.). For open-ended staging questions, predictions are strictly considered correct only when they exhibit exact lexical matches with the ground truth answers.

\subsection{Implementation Details}
In the pretrain stage, we freeze the LLM and train the slide encoder for 1 epoch with a learning rate of 1e-3. In fine-tuning, we apply LoRA to the LLM and jointly optimize both the slide encoder and LLM for 9 epochs with a learning rate of 2e-5. We use Vicuna-7b-v1.5 as the LLM, with all patches encoded into 1024-dim features via the UNI encoder. Experiments were run on an NVIDIA A100 40G GPU with Pytorch.

\subsection{Comparisons}
We compared the performance of the proposed CTIS-QA model against the baseline WSI-VQA model \cite{chen2024wsi}. We also evaluated the performance of general-purpose MLLM GPT-4o \cite{yang2023dawn} and patch-level MLLM Quilt-LLaVA \cite{seyfioglu2024quilt}. For models with input size limitations, we resized the WSIs to 1024 × 1024 pixels to accommodate their processing constraints. As shown in Table I, the CTIS-QA method demonstrates consistent improvements over existing approaches across both general and clinically curated benchmarks, reflecting its enhanced capability in understanding Whole Slide Images for visual question answering.  The performance gain is particularly evident in clinically relevant diagnostic categories, where accurate interpretation requires both fine-grained morphological analysis and integration of global architectural context.  This advancement stems from the synergistic design of our model and dataset.  By grounding the VQA task in a pathologist-informed Clinical Pathology Report Template, our benchmark avoids the pitfalls of automatically generated questions—such as hallucinations, information leakage, and non-visual reasoning—ensuring that model performance genuinely reflects visual comprehension.  Furthermore, the proposed model architecture emulates the diagnostic workflow of pathologists through a dual-stream visual encoder: the global stream captures high-level tissue organization via clustering-based feature aggregation, providing contextual awareness across the entire slide, while the local stream leverages a learnable query mechanism to selectively focus on diagnostically salient regions, preserving critical microscopic details.  The fusion of these complementary representations enables a more holistic and nuanced understanding of complex histological patterns.  Unlike models that rely solely on patch-level features or generic vision-language alignment, our approach aligns visual semantics with structured clinical knowledge, facilitating more accurate, reliable, and interpretable responses. 

We examine our model's capabilities in handling slide-level classification tasks. As shown in Table II, our method achieves superior performance on both tumor subtyping and PR status prediction, demonstrating strong slide-level reasoning abilities. The results indicate that the model effectively leverages both global architectural patterns and local morphological details for accurate classification. Particularly noteworthy are the high recall and F1 scores in subtyping, suggesting that the model is highly sensitive in identifying relevant histological subtypes while minimizing false negatives. Similarly, the balanced precision and recall in PR prediction reflect robust feature learning and generalization capabilities, outperforming existing pooling strategies and transformer-based approaches. These improvements can be attributed to our dual-stream architecture, which integrates comprehensive slide-level context through clustering-based global representation while simultaneously capturing diagnostically critical local regions via attention-guided patch selection.

\begin{table}[!h]
\centering
\caption{Performance comparison on subtyping and PR tasks. Precision (P), Recall (R), and F1-score are reported for each task.}
\renewcommand{\arraystretch}{1.2}
\resizebox{\columnwidth}{!}{%
\begin{tabular}{lcccccc}
\hline
\textbf{Method} & \multicolumn{3}{c}{\textbf{subtyping}} & \multicolumn{3}{c}{\textbf{PR}} \\
\hline
 & P & R & F1 & P & R & F1 \\
\hline
meanpooling & 0.899 & 0.940 & 0.919 & 0.796 & 0.839 & 0.817 \\
maxpooling & 0.842 & 0.955 & 0.895 & 0.742 & \textbf{0.928} & 0.825 \\
TransMIL & \textbf{0.983} & 0.895 & 0.937 & 0.821 & 0.821 & 0.821 \\
Quilt-LLaVA & 0.861 & 0.925 & 0.892 & 0.687 & 0.589 & 0.634 \\
WSI-VQA & 0.907 & 0.880 & 0.893 & 0.851 & 0.828 & 0.836 \\
\hline
CTIS-QA(Ours) & 0.942 & \textbf{0.970} & \textbf{0.955} & \textbf{0.859} & 0.875 & \textbf{0.867} \\
\hline
\end{tabular}%
}
\label{tab:subtyping_pr_results}
\end{table}

\begin{table*}[htbp]
\centering
\renewcommand{\arraystretch}{1.2}
\caption{Comparison of different settings on WSI-VQA and CTIS-Bench. BLEU, METEOR, RougeL, and Accuracy are reported for WSI-VQA; category-wise scores and average for CTIS-Bench. Pre. indicates CTIS-Align pretrain.The gray-highlighted row corresponds to pretraining directly on original pathology reports.}
\label{tab:wsi_slide_understand_results}
\resizebox{\textwidth}{!}{%
\begin{tabular}{cc|ccccc|ccccccc}
\hline
 & & \multicolumn{5}{c|}{\textbf{WSI-VQA}} & \multicolumn{7}{c}{\textbf{CTIS-Bench}} \\
\cline{3-14}
\textbf{Pre.} & \textbf{PPM} & BLEU1 & BLEU4 & METEOR & RougeL & Acc 
 & H.F. & L.C. & C.P.F. & Subtypes & Staging & M.M. & Avg \\
\hline
 &  & 0.703 & 0.651 & 0.421 & 0.638 & 0.562 & 0.464 & 0.515 & 0.591 & 0.737 & 0.547 & 0.554 & 0.568 \\
\text{\checkmark} &  & 0.759 & 0.679 & 0.471 & 0.694 & 0.638 & 0.480 & 0.576 & \textbf{0.615} & 0.825 & \textbf{0.573} & 0.583 & 0.609 \\

\text{\checkmark} & \text{\checkmark} & \textbf{0.767} & \textbf{0.701} & \textbf{0.479} & \textbf{0.701} & \textbf{0.649} & \textbf{0.523} & \textbf{0.635} & 0.599 & \textbf{0.851} & 0.541 & \textbf{0.629} & \textbf{0.630} \\
\hline
\rowcolor{gray!20}
\text{\checkmark}  & \text{\checkmark}  & 0.730 & 0.636 & 0.444 & 0.662 & 0.592 & 0.331 & 0.451 & 0.500 & 0.500 & 0.518 & 0.417 & 0.453 \\
\hline
\end{tabular}%
}

\end{table*}

\subsection{Ablation Study}

We evaluate the effectiveness of our proposed components—\textbf{CTIS-Align pretrain} and the \textbf{Patch Perception Module (PPM)}—on both the WSI-VQA and CTIS-Bench datasets. The results reveal a clear performance progression when these components are incorporated, highlighting their complementary roles in enhancing WSI understanding. The introduction of CTIS-Align pretrain significantly improves the model’s ability to align global slide-level visual content with structured clinical language, enabling more accurate semantic grounding. When combined with the PPM, which captures fine-grained local morphological details through learnable query-based attention, the model achieves further gains, demonstrating that the integration of both global context and local saliency is essential for comprehensive visual comprehension. This synergistic design allows the model to simultaneously reason about tissue architecture and microscopic diagnostic features, leading to more robust and accurate responses in complex VQA tasks. In contrast, when only original pathology report pretrain is used without the structured template and local refinement, performance drops substantially, indicating that naive pretrain on raw report text introduces noise and weak alignment, ultimately undermining diagnostic reliability. Thus, our approach emphasizes the importance of clinically informed, structured pretrain paired with a dual-stream perception mechanism for effective pathology vision-language modeling.

\section{Conclusion}

In this work, we addressed critical limitations in existing WSI-VQA datasets by introducing a clinical diagnosis template-based pipeline that systematically collects and structures pathological information. Through collaboration with pathologists and adherence to CAP Cancer Protocols, we developed the Clinical Pathology Report Template (CPRT) and constructed CTIS-Bench, a rigorously curated VQA benchmark comprising 977 WSIs and 14,879 question-answer pairs that emphasizes clinically grounded questions reflecting real diagnostic workflows. We further proposed CTIS-QA, a WSI-specific visual question answering model featuring a dual-stream architecture that captures both global slide-level context and salient local regions. Extensive experiments demonstrated that CTIS-QA consistently outperforms existing state-of-the-art models across multiple evaluation metrics, particularly excelling in complex subtype classification tasks with significant clinical value.

Our contributions establish a new paradigm for creating high-quality, clinically relevant WSI-VQA datasets while demonstrating the effectiveness of template-guided approaches in multimodal pathological analysis. Looking forward, we envision extending our template-based approach to other cancer types and pathological domains, enabling comprehensive multi-organ diagnostic systems. 
Additionally, integrating temporal pathological data and treatment response information could further enhance clinical utility, ultimately supporting more personalized and precise pathological diagnosis in clinical practice.

% \section*{Acknowledgment}

% The preferred spelling of the word ``acknowledgment'' in America is without 
% an ``e'' after the ``g''. Avoid the stilted expression ``one of us (R. B. 
% G.) thanks $\ldots$''. Instead, try ``R. B. G. thanks$\ldots$''. Put sponsor 
% acknowledgments in the unnumbered footnote on the first page.

% \section*{References}

% Please number citations consecutively within brackets \cite{b1}. The 
% sentence punctuation follows the bracket \cite{b2}. Refer simply to the reference 
% number, as in \cite{b3}---do not use ``Ref. \cite{b3}'' or ``reference \cite{b3}'' except at 
% the beginning of a sentence: ``Reference \cite{b3} was the first $\ldots$''

% Number footnotes separately in superscripts. Place the actual footnote at 
% the bottom of the column in which it was cited. Do not put footnotes in the 
% abstract or reference list. Use letters for table footnotes.

% Unless there are six authors or more give all authors' names; do not use 
% ``et al.''. Papers that have not been published, even if they have been 
% submitted for publication, should be cited as ``unpublished'' \cite{b4}. Papers 
% that have been accepted for publication should be cited as ``in press'' \cite{b5}. 
% Capitalize only the first word in a paper title, except for proper nouns and 
% element symbols.

% For papers published in translation journals, please give the English 
% citation first, followed by the original foreign-language citation \cite{b6}.

\bibliographystyle{IEEEtran.bst}
\bibliography{ref}
% \begin{thebibliography}{00}
% \bibitem{b1} G. Eason, B. Noble, and I. N. Sneddon, ``On certain integrals of Lipschitz-Hankel type involving products of Bessel functions,'' Phil. Trans. Roy. Soc. London, vol. A247, pp. 529--551, April 1955.
% \bibitem{b2} J. Clerk Maxwell, A Treatise on Electricity and Magnetism, 3rd ed., vol. 2. Oxford: Clarendon, 1892, pp.68--73.
% \bibitem{b3} I. S. Jacobs and C. P. Bean, ``Fine particles, thin films and exchange anisotropy,'' in Magnetism, vol. III, G. T. Rado and H. Suhl, Eds. New York: Academic, 1963, pp. 271--350.
% \bibitem{b4} K. Elissa, ``Title of paper if known,'' unpublished.
% \bibitem{b5} R. Nicole, ``Title of paper with only first word capitalized,'' J. Name Stand. Abbrev., in press.
% \bibitem{b6} Y. Yorozu, M. Hirano, K. Oka, and Y. Tagawa, ``Electron spectroscopy studies on magneto-optical media and plastic substrate interface,'' IEEE Transl. J. Magn. Japan, vol. 2, pp. 740--741, August 1987 [Digests 9th Annual Conf. Magnetics Japan, p. 301, 1982].
% \bibitem{b7} M. Young, The Technical Writer's Handbook. Mill Valley, CA: University Science, 1989.
% \end{thebibliography}
% \vspace{12pt}
% \color{red}
% IEEE conference templates contain guidance text for composing and formatting conference papers. Please ensure that all template text is removed from your conference paper prior to submission to the conference. Failure to remove the template text from your paper may result in your paper not being published.

\end{document}